\documentclass[conference]{IEEEtran}
\IEEEoverridecommandlockouts
\usepackage{cite}
\usepackage{amsmath,amssymb,amsfonts}
\usepackage{graphicx}
\usepackage{textcomp}
\usepackage{xcolor}
\usepackage{booktabs}
\usepackage{bm}
\usepackage{hyperref}
\usepackage{tabularx}

\def\BibTeX{{\rm B\kern-.05em{\sc i\kern-.025em b}\kern-.08em
    T\kern-.1667em\lower.7ex\hbox{E}\kern-.125emX}}
\begin{document}

\title{UDAV: Uncertainty-Driven Adaptive VLM Waypoint Planner}

\author{Ghazal Farhani$^{*}$ and Shabnam Shabani%
\thanks{The authors are with the Automotive and Surface Transportation
Research Centre, National Research Council Canada, London, Ontario, Canada.}%
\thanks{$^{*}$Corresponding author: 
(\texttt{ghazal.farhani@nrc-cnrc.gc.ca}).}%
}

\maketitle

\begin{abstract}

A key challenge in using vision–language models (VLMs) for aerial off-road navigation is that their predicted routes provide no indication of reliability. We propose UDAV, an \emph{Uncertainty-Driven Adaptive VLM Waypoint Planner} for UAV-guided UGV navigation. UDAV generates repeated stochastic trajectories from the VLM, selects their medoid as the nominal self-consistent route, and uses their spatial dispersion to estimate predictive uncertainty. If the maximum uncertainty among interior waypoints exceeds a threshold, the planner invokes an additional reconsideration stage; otherwise, it returns the medoid directly.

Across 400 held-out trajectory queries from two UAV flights, stochastic medoid selection reduces mean ADE from 147.4~px for a deterministic VLM prediction to 115.9~px. The complete UDAV planner achieves a mean ADE of 110.4~px, a 25.1\% reduction, with valid trajectories for all 400 queries. UDAV also achieves the lowest P90 and P95 errors among all evaluated configurations, including a higher-budget $K=10$ consensus baseline. Relative to the $K=5$ medoid, it reduces P90/P95 ADE from 225.3/326.0~px to 199.0/290.8~px. Thus, stochastic VLM predictions provide both a stronger nominal route and an actionable uncertainty signal for selectively mitigating large planning errors.

\end{abstract}


\begin{figure*}[!t]
    \centering

    \begin{minipage}[t]{0.78\textwidth}
        \vspace{0pt}
        \centering
        \includegraphics[width=\linewidth]{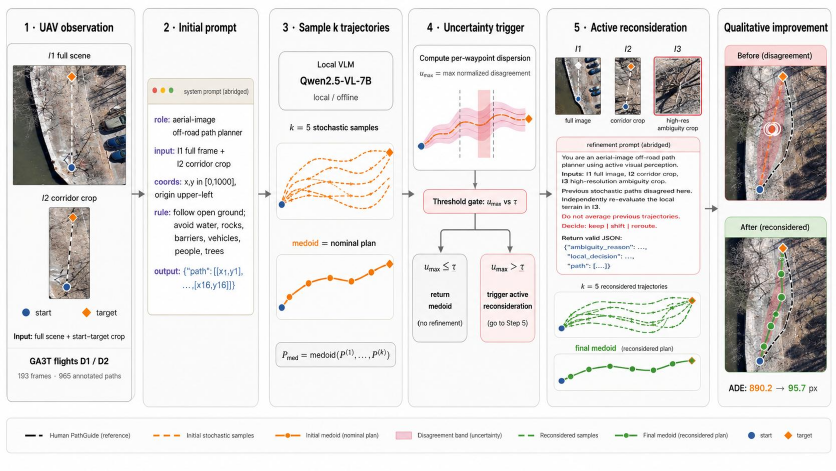}
    \end{minipage}
    \hfill
    \begin{minipage}[t]{0.21\textwidth}
        \vspace{0pt}

        \noindent\fbox{%
        \begin{minipage}{\dimexpr\linewidth-2\fboxsep-2\fboxrule\relax}
        \scriptsize
        \textbf{Prompt 1: Initial planning (abridged)}\\[2pt]
        Given the full aerial image $I_1$ and a higher-resolution
        start--target corridor crop $I_2$, generate a traversable
        off-road path in Image~1 coordinates. Follow open ground and
        avoid water, barriers, large rocks, vehicles, people, animals,
        trees, benches, and other unsafe obstacles. Return only valid
        JSON containing exactly 16 ordered waypoints.
        \end{minipage}%
        }

        \vspace{6pt}

        \noindent\fbox{%
        \begin{minipage}{\dimexpr\linewidth-2\fboxsep-2\fboxrule\relax}
        \scriptsize
        \textbf{Prompt 2: Refinement (abridged)}\\[2pt]
        Given the full image $I_1$, corridor crop $I_2$, and
        high-resolution disagreement crop $I_3$, independently
        re-evaluate the ambiguous terrain. Do not average or imitate
        the previous trajectories. Decide whether to \textbf{keep},
        \textbf{shift}, or \textbf{reroute}, and return exactly
        16 ordered waypoints in valid JSON.
        \end{minipage}%
        }
    \end{minipage}

    \caption{Schematic of the proposed UDAV planner, with the initial
    planning and uncertainty-triggered refinement prompts shown in
    abridged form. The complete prompts used in the experiments are included
    in the released reproducibility package.}
    \label{fig:schematic}
\end{figure*}

\section{Introduction}
\label{sec:introduction}

Autonomous navigation in unstructured off-road environments, with applications in search and
rescue, disaster response, and remote inspection
\cite{chung2023into}, has remained a fundamental challenge in field robotics. Unlike structured roads or indoor
environments, off-road scenes lack explicit traversability boundaries, and
the navigability of visually ambiguous terrain often cannot be inferred from
geometry alone. Unmanned aerial vehicle (UAV)--unmanned ground vehicle (UGV)
collaboration can address these challenges. The UGV keeps local sensing
and control, while a UAV can quickly acquire a global view of the
environment, spot hazards beyond the UGV's field of view, and provide
high-level route guidance \cite{liu2022review,cladera2024enabling}.

In a classical aerial-planning pipeline, first semantic or
traversability information is inferred, converted into a cost map, and then a
route with a classical planner such as A$^{*}$
\cite{hosseinpoor2021traversability,hart1968formal} is proposed. This works well when
terrain classes and traversal costs are known in advance, but it requires a
predefined semantic class set, an explicit class-to-cost mapping, and
environment-specific supervision, restrictive assumptions in previously
unseen or highly variable off-road terrain.

Vision--language models (VLMs) offer an alternative as they associate visual
observations with semantic and spatial concepts and can directly produce
structured navigation outputs \cite{jia2026omnispatial,cladera2025air}. Recent work has begun applying them to embodied navigation
\cite{liu2026cofl,chen2026worldmap}. However, aerially guided off-road planning with
locally deployable VLMs, remains largely unexplored, especially
when the planner must judge the reliability of a predicted trajectory before
execution.

A single deterministic VLM prediction says nothing about its own reliability.
Under stochastic decoding, the same scene can produce substantially different
waypoint sequences, but a deterministic prediction hides this variability and
offers no basis for deciding whether additional reasoning is warranted. In this study, we
ask whether disagreement among stochastic trajectory predictions can serve as
an actionable signal for when an otherwise self-consistent plan should be
reconsidered.

To answer this, we develop \textbf{UDAV}, an uncertainty-driven adaptive VLM
waypoint planner for UAV-guided off-road navigation. A locally deployable VLM
is fine-tuned to map an aerial image, a start location, and a target location
to an ordered sequence of $N$ image-space waypoints. At inference time the
planner draws $K$ stochastic trajectory predictions and uses them twice:
their medoid becomes the nominal self-consistent trajectory, and the
per-waypoint covariance across the samples yields a geometric estimate of
predictive uncertainty. Consensus and uncertainty thus come from the same
samples, with no separate uncertainty-estimation model. If the maximum
normalized uncertainty stays below a predefined threshold, UDAV returns the
stochastic medoid directly; if it is exceeded, the model is told that its
previous predictions disagreed within the identified region and is prompted
to reconsider the route, and the final trajectory is taken as the medoid of
the reconsidered predictions. 

Two empirical observations motivate this design. Stochastic trajectory
disagreement is significantly associated with trajectory error, and the
benefit of stochastic sampling extends beyond uncertainty estimation:
medoid-based self-consistency alone substantially improves accuracy over a
single deterministic prediction. In UDAV, uncertainty consequently plays a
distinct role, not defining the nominal trajectory, but deciding when an
already self-consistent plan warrants a second planning pass.

We evaluate UDAV on two flight sequences from the GA3T UAV--UGV off-road
dataset \cite{cai2026ga3t}: 193 spatially separated aerial observations with
five candidate UGV paths each, giving 965 annotated path examples, of which a
fixed held-out test set of 400 trajectory queries is used for evaluation.
Deterministic VLM planning reaches a mean average displacement error (ADE) of
147.4~px on this test set; selecting the $K=5$ stochastic medoid brings this
down to 115.9~px, and the full UDAV planner at $\tau=0.6$ to 110.4~px, a
25.1\% reduction relative to deterministic planning, with valid trajectories
for all 400 queries. On the 174 queries where reconsideration is activated,
mean ADE drops from 153.7~px to 141.0~px, and over the full test set UDAV
attains the lowest P90 and P95 errors of all evaluated configurations,
below even a higher-budget $K=10$ consensus baseline.

This work makes three contributions (the overall planner is shown in
Fig.~\ref{fig:schematic}). First, we formulate stochastic VLM trajectory
disagreement as a geometric predictive-uncertainty signal for aerially guided
off-road waypoint planning. Second, we introduce an adaptive planning
mechanism that uses medoid self-consistency as the nominal prediction and
invokes a second VLM planning pass only when trajectory disagreement exceeds
a prescribed threshold. Third, we show that stochastic sampling provides two
complementary benefits: medoid self-consistency improves nominal trajectory
accuracy, while uncertainty-triggered reconsideration reduces large-error
failures. To facilitate reproducibility, we release the ground-truth
annotations used for fine-tuning and evaluation, the final fine-tuned model
weights, and detailed data-preparation and reproduction instructions in the
accompanying
\href{https://drive.google.com/drive/folders/1hnLkDzkfL90dPkvioRoLQ40c0fL1dHsI}{reproducibility package}.

\section{Related Work}

\subsection{Vision-Language Models for Navigation}

Because VLMs emit text while control requires continuous coordinates, a
recurring strategy is to render candidate actions in the image and let the
model choose among them. \textit{Nasiriany et al.}~\cite{nasiriany2024pivot}
formalise this as PIVOT, which casts spatial reasoning as iterative visual
question answering: candidates are drawn on the image, the VLM selects the
most promising, and a refined distribution is resampled and requeried until
the answer converges. \textit{Sathyamoorthy et al.}~\cite{sathyamoorthy2024convoi}
apply the same principle to navigation in CoNVOI, overlaying numbered
markers on obstacle-free regions of a ground-level RGB image and prompting
cloud-hosted VLMs (GPT-4V, Gemini) to select the markers satisfying a
context-dependent behaviour phrase. Their ablation is instructive: removing
the markers raises the proportion of unacceptable paths from $0\%$ to
$85.7\%$, indicating that current VLMs cannot reliably reason about free
space without an explicit spatial prior. Both methods consequently narrow
an externally supplied proposal set rather than generating a route
directly, and the authors of CoNVOI identify the absence of a global
top-down view as a limitation to be addressed in future work.

A different composition is used by \textit{Shah et al.}~\cite{shah2023lm}
in LM-Nav, which combines three pre-trained models without fine-tuning: an
LLM parses the instruction into landmark names, CLIP grounds those names in
observations, and a goal-conditioned policy (ViNG) executes the motion
between them. Language thus determines \emph{which} landmarks to visit,
while the route between them is produced by a separate visual policy.
LM-Nav also depends on a pre-collected traversal graph, and therefore does
not operate in previously unobserved terrain.

\subsection{Traversability Estimation for Off-Road Navigation}

Classical off-road pipelines infer terrain properties and convert them into
a planning cost. \textit{Guan et al.}~\cite{guan2022ga} proposed GA-Nav,
which segments unstructured outdoor scenes into coarse traversability
groups and assigns each group a traversal cost, producing smoother routes
than pixel-wise semantic segmentation.
\textit{Shaban et al.}~\cite{shaban2022semantic} similarly classify terrain
into predefined semantic classes for off-road driving. Both require the
class set and the class-to-cost mapping to be specified in advance, and
both are specific to the platform and the environment. An alternative line
learns traversability from experience rather than from labels: WayFAST
\cite{gasparino2022wayfast} supervises a traversability predictor using
traction estimates recorded during driving, which removes the need for
hand-defined costs but presumes prior traversal of representative terrain.

\subsection{UAV--UGV Collaboration}

Aerial--ground teaming has been studied extensively, spanning
inter-vehicle communication, task allocation, and navigation optimisation;
\cite{munasinghe2024comprehensive} provides a comprehensive review. Within
this body of work, however, the use of VLMs for aerially guided ground
navigation remains largely unexplored. Most closely related to our setting,
\textit{Cladera et al.}~\cite{cladera2025air} demonstrate that decomposing
an air--ground mission into interdependent sub-tasks, relayed between the
UAV and the UGV, supports effective autonomous operation. Their emphasis is
on coordinating and sequencing these sub-tasks; by contrast, we task the
language model with producing the waypoint sequence that carries the UGV to
an assigned target, together with an explanation of the visual evidence
supporting it.

\section{Method}
\label{sec:uncertainty_waypoint_planning}

UDAV uses stochastic VLM predictions: their medoid provides a self-consistent nominal trajectory, while their
geometric dispersion determines whether additional visual reasoning is
required. The overall procedure is illustrated in
Fig.~\ref{fig:schematic}.

A trajectory is represented by $N=16$ ordered image-space waypoints,
$W=(w_1,\ldots,w_N)$, where $w_i=[x_i,y_i]^\top$. Given the full aerial
image $I_1$, corridor crop $I_2$, start $s$, and target $t$, the
fine-tuned VLM generates $K$ stochastic trajectories,
\begin{equation}
    W^{(k)}
    \sim
    \pi_\theta(W\mid I_1,I_2,s,t),
    \qquad k=1,\ldots,K,
\end{equation}
with $K=5$ in our experiments.

\subsection{Stochastic Consensus}

Trajectory similarity is measured as
\begin{equation}
    D_T(W^{(a)},W^{(b)})
    =
    \frac{1}{N}
    \sum_{i=1}^{N}
    \left\|w_i^{(a)}-w_i^{(b)}\right\|_2 .
    \label{eq:trajectory_distance}
\end{equation}
The nominal trajectory is selected as the medoid,
\begin{equation}
\begin{aligned}
W^{\mathrm{med}} &= W^{(k^*)},\\
k^* &=
\operatorname*{arg\,min}_{k}
\frac{1}{K-1}
\sum_{\substack{j=1\\j\neq k}}^{K}
D_T\!\left(W^{(k)},W^{(j)}\right).
\end{aligned}
\label{eq:initial_medoid}
\end{equation}
Unlike pointwise averaging, the medoid remains one of the trajectories
generated by the VLM.

\subsection{Uncertainty-Triggered Reconsideration}

For each interior waypoint, the empirical covariance across stochastic predictions is
\begin{equation}
    \bm{\Sigma}_i =
    \frac{1}{K-1}\sum_{k=1}^{K}
    \left(w_i^{(k)}-\bm{\mu}_i\right)
    \left(w_i^{(k)}-\bm{\mu}_i\right)^\top ,
\end{equation}
where $\bm{\mu}_i$ is the corresponding sample mean. The normalized
uncertainty is then defined as
\begin{equation}
    u_i =
    \frac{
    \sqrt{\chi^2_{2,0.95}\lambda_{\max}(\bm{\Sigma}_i)}
    }{\|t-s\|_2},
    \qquad i=2,\ldots,N-1 .
    \label{eq:normalized_uncertainty}
\end{equation}

If $\max_i u_i\leq\tau$, the nominal medoid is returned directly.
Otherwise, uncertain waypoints are grouped into maximal contiguous
regions, and the region containing the largest uncertainty is selected.
Optionally, a high-resolution crop $I_3$ of this region is supplied to
the VLM together with $I_1$ and $I_2$; the core method operates without
this additional crop, and its effect is evaluated separately in
Sec.~\ref{subsec:visual_ablation}. The refinement prompt informs the
model that its previous predictions disagreed and asks it to
independently reassess the local terrain rather than averaging the
previous trajectories (the right panel of Fig.~\ref{fig:schematic} shows
the prompt).

The model generates $K_{\mathrm{ref}}=5$ reconsidered trajectories,
\begin{equation}
    \widetilde{W}^{(q)}
    \sim
    \pi_\theta
    (W\mid I_1,I_2,I_3,s,t,\mathcal{C}_\tau),
    \qquad q=1,\ldots,K_{\mathrm{ref}},
\end{equation}
where $\mathcal{C}_\tau$ denotes the uncertainty cue and $I_3$ is
included only in the crop-augmented variant. Their medoid,
$W^{\mathrm{ref}}$, is selected using the same trajectory-distance
criterion in Eq.~\eqref{eq:trajectory_distance}. The final planner is
therefore
\begin{equation}
    W^{\mathrm{final}}
    =
    \begin{cases}
        W^{\mathrm{med}}, & \max_i u_i\leq\tau,\\[1mm]
        W^{\mathrm{ref}}, & \max_i u_i>\tau.
    \end{cases}
    \label{eq:selective_replanning}
\end{equation}

\subsection{Dataset Overview}\label{subsec:data}
We used two datasets from the GA3T repository:
\texttt{20260318-140904-dataset}, referred to as \textbf{D1}, and
\texttt{20260318-141807-dataset}, referred to as \textbf{D2} \cite{cai2026ga3t}.
Because both the UAV and UGV move relatively slowly, consecutive UAV frames are often highly similar and provide limited additional information for VLM-based path learning. We therefore subsampled each sequence to retain frames that were visually distinct while still preserving the progression of the UGV along the trail.  

D1 originally contained 790 UAV frames, from which 99 representative frames were selected. For each selected frame, five human-annotated \textbf{PathGuide trajectories} were generated. One trajectory represented the primary forward direction of travel along the trail and was used to preserve correspondence with the route subsequently followed by the UGV. The remaining four trajectories represented alternative feasible paths through the same scene, increasing the diversity of the fine-tuning data and exposing the model to more challenging navigation alternatives. This resulted in a total of 495 trajectory samples. D1 was partitioned at the frame level into 69 training frames (345 trajectories) and 30 permanently held-out test frames (150 trajectories). To reduce information leakage arising from temporally adjacent and visually similar UAV observations, consecutive frames were first grouped into blocks of 10, and the train--test partition was performed at the block level. Consequently, all five PathGuide trajectories associated with a given frame remained within the same partition.

The same frame-selection strategy was applied to D2. From the original 850 UAV frames, 94 representative frames were retained, producing 470 PathGuide trajectory samples. A fixed set of 50 complete frames, corresponding to 250 trajectories, was reserved as the permanent D2 test set, while the remaining 44 frames (220 trajectories) were used exclusively for domain adaptation. 

In addition to the PathGuide trajectories, scene objects and relevant terrain features in each selected frame (including rocks, water, traversable path regions, vehicles, and other obstacles) were manually annotated using \textbf{LabelMe}. These annotations served two purposes: they provided supervision for fine-tuning the semantic segmentation model, and they were used to construct the obstacle-aware cost maps required by the A$^{*}$ planner, which is used as a classical method for comparison against our results. The resulting trajectory annotations are available in the \href{https://drive.google.com/drive/folders/1hnLkDzkfL90dPkvioRoLQ40c0fL1dHsI}{PathGuide annotation dataset}.

\subsubsection{VLM model}

Initially, without any fine-tuning, we evaluated four VLMs on a subset of the test data: Qwen2.5-VL-7B, InternVL3-8B, LLaVA-OneVision-7B, and SmolVLM2-2.2B. Among these models, Qwen2.5-VL-7B and InternVL3-8B achieved the lowest waypoint prediction error compared with the ground-truth trajectories. In preliminary fine-tuning experiments, Qwen2.5-VL-7B adapted substantially more effectively than InternVL3-8B; all subsequent experiments therefore use Qwen2.5-VL-7B.
A further advantage of Qwen2.5-VL-7B is that it can be deployed offline, making it particularly useful in environments with limited or no internet connectivity. The prompt is a simple text instruction asking the model to use the visual information to determine the best path from the UGV to the target and generate 16 waypoints along that path.

\subsection{A$^{*}$ Planning}

A$^{*}$~\cite{hart1968formal} is a graph-search algorithm that recovers a
minimum-cost path between a start and a goal node while excluding regions
deemed non-traversable, and it remains widely used in ground- and
aerial-robot planning. In grid-based implementations the environment is
encoded as a cost map in which each cell carries a traversal cost derived
from obstacle occupancy, terrain type, and a desired clearance margin around
hazardous regions. The recovered path therefore reflects the cost map as much
as the search itself.

We fine-tune \texttt{segformer-b5}~\cite{xie2021segformer} on the
113 non-test training/adaptation frames (69 from D1 and 44 from D2) over nine
classes: background, water, barrier, vehicle, person, animal, rock, bench, and
tree. The resulting semantic predictions are converted into the obstacle-aware
cost map used by the A$^{*}$ planner. No permanent-test annotation is used to
train the segmentation model.

\subsection{Metrics of Comparison}

We evaluate geometric similarity between the predicted and reference
trajectories using four complementary image-space metrics:
\begin{itemize}
    \item \textbf{ADE (Average Displacement Error):} the mean Euclidean
    distance between corresponding points of the resampled predicted and
    reference trajectories.

    \item \textbf{MidADE:} ADE computed over the interior waypoints only,
    excluding the fixed start and target endpoints.

    \item \textbf{Chamfer Distance:} a symmetric nearest-neighbor distance
    between the two trajectory point sets.

    \item \textbf{Hausdorff Distance:} the maximum nearest-neighbor mismatch
    between the two trajectory point sets.
\end{itemize}
Lower values indicate greater geometric similarity.

\subsection{Fine-Tuning Setup}
The model is first adapted on the D1 training partition, which
contains 70\% of the selected D1 samples, using QLoRA with 4-bit NF4
quantization and double quantization. Training is performed for three epochs
with paged 8-bit AdamW, a cosine learning-rate schedule, an initial learning
rate of $2\times10^{-4}$, and an effective batch size of eight. Computation is
carried out in BF16. As described in Sec.~\ref{subsec:data}, the training and
test samples are spatially separated to reduce leakage between temporally and
spatially adjacent UAV observations. The resulting D1-adapted model is
evaluated on both the held-out D1 test set and D2, which corresponds to a
different UAV flight and environment. The model is then further adapted on D2
for one epoch using a reduced learning rate of $5\times10^{-5}$ and reevaluated
on both test sets. All runs are performed in Google Colab on an NVIDIA A100
80~GB GPU, using BF16 computation and a maximum of 768 visual tokens per
image.

% ============================================================
% RESULTS
% ============================================================

\section{Results}
\label{sec:results}

% ------------------------------------------------------------
\subsection{Sequential Fine-Tuning and Cross-Flight Generalization}
\label{subsec:Fine-tune}

Table~\ref{tab:qwen_finetuning_comparison} summarizes the trajectory
prediction performance of Qwen2.5-VL-7B before and after sequential
fine-tuning.

Fine-tuning on D1 substantially improves performance relative to the
zero-shot model. On D1, ADE decreases from 241.8~px to 149.3~px, while
Chamfer distance decreases from 140.0~px to 75.8~px. Importantly, the
D1-adapted model also transfers to the unseen D2 flight, reducing ADE
from 218.8~px to 194.7~px without exposure to D2 during training.
This demonstrates that the learned waypoint representation exhibits
cross-flight generalization, although a noticeable domain gap remains.

Subsequent adaptation on D2 further reduces D2 ADE from 194.7~px to
156.8~px, corresponding to a 19.5\% improvement relative to the
D1-only model. Adaptation to D2 does not degrade performance on D1;
instead, D1 ADE further decreases from 149.3~px to 131.7~px, with
corresponding improvements in Mid-ADE, Chamfer distance, and Hausdorff
distance. Consequently, sequential D1$\rightarrow$D2 adaptation reduces
the combined ADE from 177.8~px to 147.4~px, with no evidence of
catastrophic forgetting in the evaluated setting.

\begin{table*}[t]
\centering
\caption{Trajectory prediction performance of Qwen2.5-VL-7B before and
after sequential fine-tuning.}
\label{tab:qwen_finetuning_comparison}
\small
\begin{tabularx}{\textwidth}{lXccccc}
\toprule
\textbf{Dataset} &
\textbf{Model Condition} &
\textbf{ADE (px) $\downarrow$} &
\textbf{Mid-ADE (px) $\downarrow$} &
\textbf{Chamfer (px) $\downarrow$} &
\textbf{Hausdorff (px) $\downarrow$} &
\textbf{Valid Path (\%) $\uparrow$} \\
\midrule

D1 & Qwen (zero-shot)
& 241.8 & 250.3 & 140.0 & 435.7 & 94.7 \\

D1 & Qwen after D1 fine-tuning
& 149.3 & 169.7 & 75.8 & 202.9 & 99.3 \\

D1 & Qwen after D1$\rightarrow$D2 fine-tuning
& $\mathbf{131.7}$
& $\mathbf{149.6}$
& $\mathbf{74.5}$
& $\mathbf{167.1}$
& $\mathbf{100.0}$ \\

\midrule

D2 & Qwen (zero-shot)
& 218.8 & 225.8 & 130.4 & 411.7 & 92.4 \\

D2 & Qwen after D1 fine-tuning
& 194.7 & 219.8 & 101.9 & 253.0 & $\mathbf{100.0}$ \\

D2 & Qwen after D1$\rightarrow$D2 fine-tuning
& $\mathbf{156.8}$
& $\mathbf{177.7}$
& $\mathbf{74.7}$
& $\mathbf{175.6}$
& 99.2 \\

\midrule

\textbf{Combined} & Qwen (zero-shot)
& 227.6 & 235.1 & 134.1 & 420.9 & 93.2 \\

\textbf{Combined} & Qwen after D1 fine-tuning
& 177.8 & 201.1 & 92.1 & 234.3 & $\mathbf{99.8}$ \\

\textbf{Combined} & Qwen after D1$\rightarrow$D2 fine-tuning
& $\mathbf{147.4}$
& $\mathbf{167.1}$
& $\mathbf{74.7}$
& $\mathbf{172.4}$
& 99.5 \\

\bottomrule
\end{tabularx}

\vspace{1mm}
\footnotesize
Lower ADE, Mid-ADE, Chamfer distance, and Hausdorff distance indicate
greater geometric similarity to the ground-truth trajectory.
Valid Path denotes the percentage of predictions that could be parsed
as valid trajectories.
\end{table*}

% ------------------------------------------------------------
\subsection{Stochastic Consensus and Uncertainty as a Reliability Signal}
\label{subsec:uncertainty_results}

Before evaluating the complete adaptive planner, we examine the two
quantities on which UDAV is built: stochastic consensus and predictive
disagreement.

Table~\ref{tab:qwen_uncertainty_latency} reports the computational cost
of stochastic sampling. Because multiple trajectories are generated
within the same inference call, latency grows sublinearly with the number
of samples. Increasing from $K=1$ to $K=5$ raises mean latency from
12.95~s to 15.57~s, an increase of approximately 20\%, while $K=10$
requires 18.51~s. We therefore use $K=5$ as the default stochastic
sampling budget for the uncertainty-driven planner and separately
evaluate $K=10$ as a higher-compute consensus baseline.

\begin{table}[t]
\centering
\caption{Inference latency for different numbers of stochastic samples.}
\label{tab:qwen_uncertainty_latency}
\small
\setlength{\tabcolsep}{4pt}
\begin{tabular}{ccccc}
\toprule
\textbf{$K$} &
\textbf{Mean} &
\textbf{Median} &
\textbf{Time/Path} &
\textbf{Relative} \\
\midrule
1  & 12.95 & 13.65 & 12.95 & 1.00$\times$ \\
3  & 14.64 & 15.11 & 4.88  & 1.13$\times$ \\
\textbf{5} & \textbf{15.57} & \textbf{15.91}
& \textbf{3.11} & \textbf{1.20$\times$} \\
10 & 18.51 & 19.00 & 1.85 & 1.43$\times$ \\
\bottomrule
\end{tabular}

\vspace{1mm}
\footnotesize
Latency values are in seconds. Time/Path denotes total generation time
divided by the number of sampled trajectories.
\end{table}

\begin{figure}[t]
    \centering
    \includegraphics[width=\linewidth]{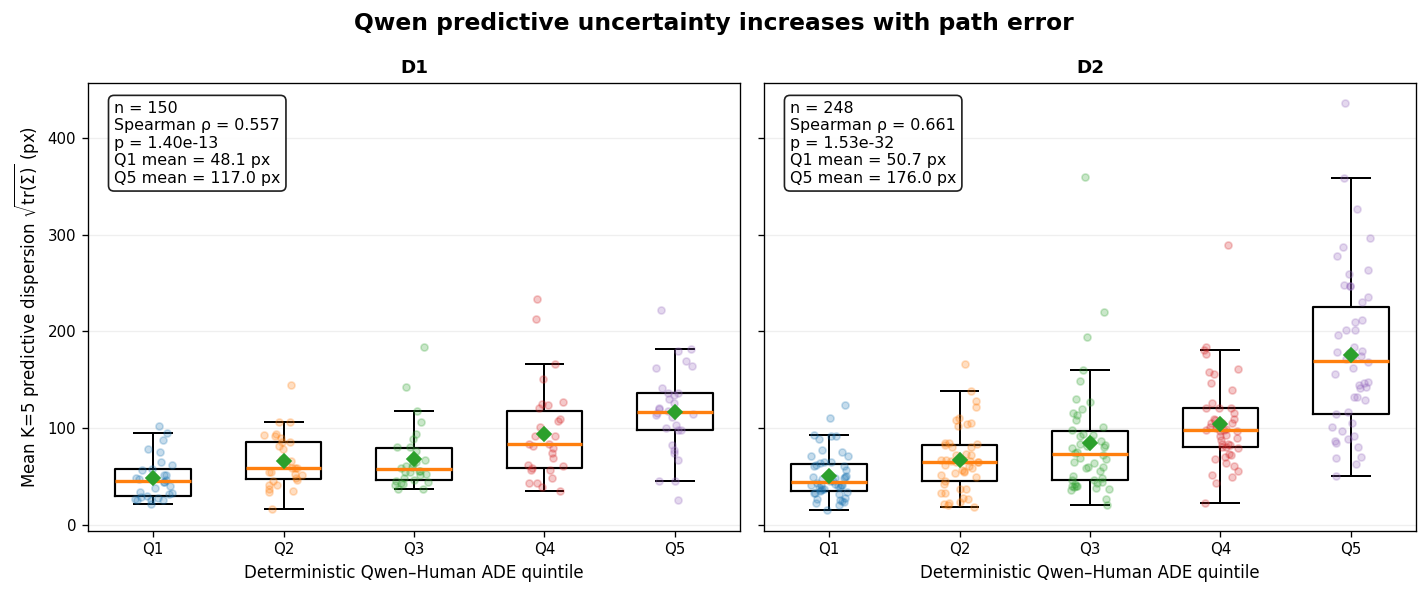}
    \caption{\textbf{Predictive disagreement is associated with trajectory
    error.} Test cases are binned by trajectory-error quintile and plotted
    against $K=5$ predictive dispersion. Dispersion increases across error
    groups in both flights (Spearman $\rho=0.557$ on D1 and $0.661$ on D2).}
    \label{fig:error_plot_box}
\end{figure}

More importantly, stochastic disagreement is informative about trajectory
reliability. Using $K=5$, stochastic trajectories are generated for the
complete permanent test sets. Figure~\ref{fig:error_plot_box} groups cases
by trajectory error and shows that predictive dispersion increases with
error. At the case level, predictive uncertainty is positively correlated
with trajectory error, with Spearman correlations of $\rho=0.557$ on D1
and $\rho=0.661$ on D2; both associations are statistically significant.

This result is central to UDAV: disagreement among repeated VLM
predictions is not merely sampling noise, but provides an online signal
of planning difficulty that can be used to decide when additional
reasoning is warranted.

% ------------------------------------------------------------
\subsection{Consensus Planning and Uncertainty-Guided Reconsideration}
\label{subsec:planner_results}

Table~\ref{tab:main_planner_comparison} summarizes the principal planning
results over all 400 permanent-test queries. A single deterministic VLM
prediction achieves a mean ADE of 147.4~px and produces valid
trajectories for 398 of the 400 queries.

Selecting the medoid of only $K=5$ stochastic predictions substantially
improves this result, reducing mean ADE from 147.4~px to 115.88~px while
producing valid trajectories for every query. This 21.4\% reduction
demonstrates that stochastic self-consistency alone provides a
substantially stronger nominal planner than a single deterministic VLM
generation.

Increasing the stochastic sampling budget to $K=10$ provides an
additional improvement, reducing mean ADE to 108.58~px and median ADE
to 80.37~px. This establishes that larger-sample consensus can further
improve nominal trajectory accuracy. However, $K=10$ requires the larger
sampling budget for every query.

UDAV instead uses the $K=5$ ensemble both to construct the nominal medoid
and to determine when additional reasoning is required. The threshold
sweep in Table~\ref{tab:main_planner_comparison} illustrates the resulting
accuracy--computation trade-off. At $\tau=0.6$, reconsideration is
activated for 174 of 400 queries (43.5\%), while the remaining 56.5\%
return the initial stochastic medoid directly.

At this operating point, UDAV achieves a mean ADE of 110.36~px, close
to the 108.58~px obtained by the $K=10$ medoid. With
$K_{\mathrm{ref}}=5$, the selective policy corresponds to an average
sampling budget of $5 + 0.435(5) = 7.18$ generated trajectories per query, compared with 10 trajectories for
the $K=10$ consensus baseline.

The distinction between the two strategies becomes particularly
interesting in the high-error tail. Although the $K=10$ medoid achieves
the lowest mean ADE, UDAV at $\tau=0.6$ achieves slightly lower P90 and
P95 errors: 198.98~px and 290.78~px, compared with 204.92~px and
292.01~px for the $K=10$ medoid. Thus, larger-sample consensus primarily
improves average trajectory accuracy, whereas uncertainty-guided
reconsideration provides a targeted mechanism for addressing difficult
cases without applying the larger inference budget uniformly.

On the 174 queries that activate reconsideration at $\tau=0.6$, the
initial $K=5$ medoid has a mean ADE of 153.71~px, compared with
141.01~px after reconsideration, corresponding to a mean reduction of
12.71~px. Reconsideration improves 50.0\% of the triggered trajectories.
The paired one-sided Wilcoxon test is not significant ($p=0.335$), and
the bootstrap confidence interval for the all-query mean improvement
includes zero. The second planning stage should therefore not be
interpreted as a guaranteed improvement for every uncertain query;
rather, its benefit is concentrated in a subset of difficult cases.

\begin{table}[t]
\centering
\caption{Full-test-set waypoint-planning performance over 400 trajectory
queries. Lower trajectory errors are better.}
\label{tab:main_planner_comparison}
\scriptsize
\setlength{\tabcolsep}{3.2pt}
\resizebox{\columnwidth}{!}{
\begin{tabular}{lcccccc}
\toprule
\textbf{Method}
& \textbf{Valid}
& \textbf{Mean ADE}
& \textbf{Median}
& \textbf{P90}
& \textbf{P95}
& \textbf{Reconsider.} \\
\midrule

Deterministic VLM
& 99.5\%
& 147.4
& 92.6
& 359.5
& --
& -- \\

$K=5$ stochastic medoid
& 100.0\%
& 115.88
& 84.06
& 225.31
& 326.04
& -- \\

$K=10$ stochastic medoid
& 100.0\%
& \textbf{108.58}
& \textbf{80.37}
& 204.92
& 292.01
& -- \\

\midrule

UDAV, $\tau=0.30$
& 100.0\%
& 115.03
& 87.92
& 227.02
& 326.89
& 71.0\% \\

UDAV, $\tau=0.50$
& 100.0\%
& 109.58
& 84.49
& 202.13
& 297.48
& 49.0\% \\

UDAV, $\tau=0.60$
& 100.0\%
& 110.36
& 83.69
& \textbf{198.98}
& \textbf{290.78}
& 43.5\% \\

UDAV, $\tau=0.70$
& 100.0\%
& 113.41
& 82.59
& 227.38
& 326.04
& 38.5\% \\

UDAV, $\tau=0.80$
& 100.0\%
& 110.32
& 83.18
& 213.90
& 322.30
& 35.8\% \\

UDAV, $\tau=1.00$
& 100.0\%
& 116.66
& 86.71
& 227.02
& 327.82
& 30.0\% \\

UDAV, $\tau=1.20$
& 100.0\%
& 116.09
& 84.06
& 225.75
& 326.04
& 26.5\% \\

\bottomrule
\end{tabular}
}

\vspace{1mm}
\footnotesize
ADE values are in pixels. ``Reconsider.'' denotes the fraction of
queries for which uncertainty exceeds $\tau$ and triggers the second
planning stage. The $K=10$ medoid achieves the lowest mean ADE, while
UDAV at $\tau=0.60$ achieves the lowest P90 and P95 errors.
\end{table}

The threshold sweep also shows that reconsidering more trajectories does
not necessarily improve performance. At $\tau=0.3$, 71.0\% of queries
are escalated, yet the resulting mean ADE is 115.03~px, only marginally
better than the $K=5$ medoid. Conversely, thresholds above approximately
$1.0$ trigger too few reconsiderations to provide a consistent benefit.
The useful operating region therefore lies between these extremes,
illustrating that the uncertainty threshold controls a meaningful
trade-off between additional computation and trajectory accuracy. We
note that $\tau$ was selected from this sweep rather than calibrated on
a separate validation split; mean ADE varies by less than 4~px across
$\tau\in[0.5,0.8]$, although the tail percentiles are more sensitive to
the threshold.

% IMPORTANT:
% Update the dashed threshold shown inside Tau-choice.png from 0.8 to 0.6
% — the caption below now states tau=0.6 but the rendered figure still
% draws the line at 0.8.

\begin{figure}[t]
    \centering
    \includegraphics[width=0.95\linewidth]{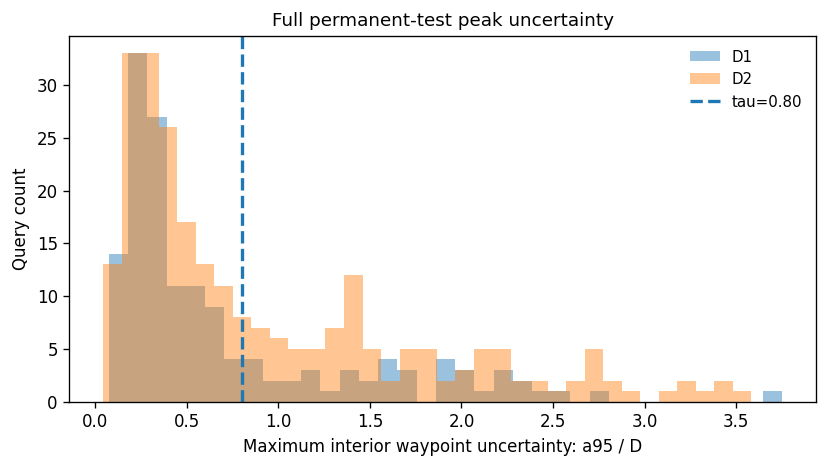}
    \caption{Distribution of the maximum normalized interior-waypoint
    uncertainty $u_{\max}$ over the D1 and D2 permanent test sets.
    At $\tau=0.6$, 174 of 400 queries (43.5\%) trigger reconsideration.}
    \label{fig:tau-choice}
\end{figure}

\begin{figure}[t]
    \centering
    \includegraphics[width=1\linewidth]{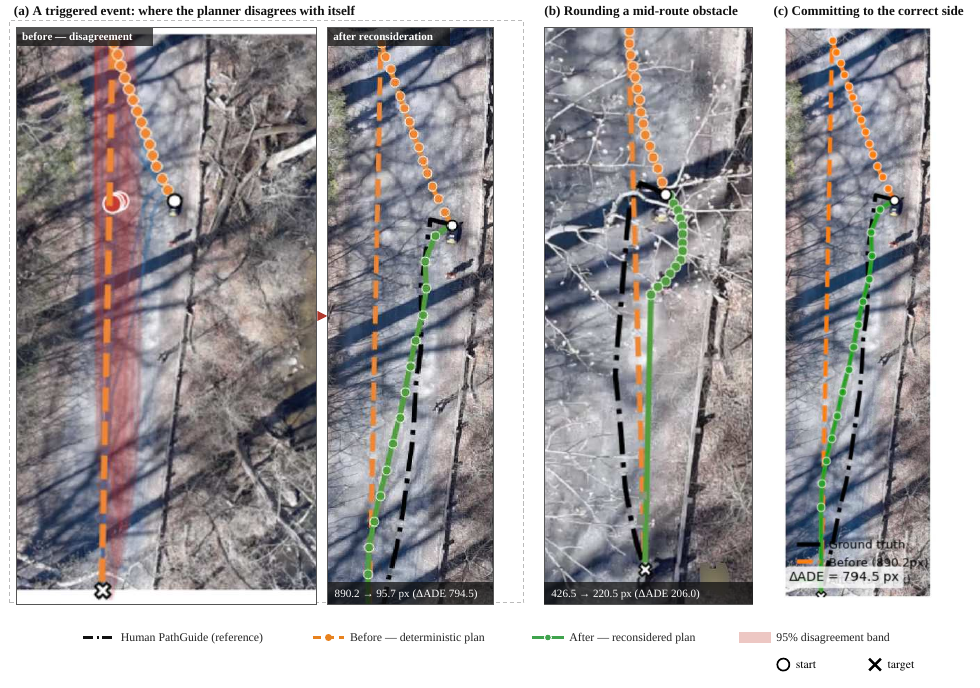}
    \caption{\textbf{Qualitative examples of uncertainty-triggered
    reconsideration.} Stochastic trajectory disagreement identifies
    difficult regions in which an additional planning pass can modify
    the route toward the human-annotated reference.}
    \label{fig:refinement_result}
\end{figure}

Qualitative examples in Fig.~\ref{fig:refinement_result} illustrate the
behavior motivating the adaptive stage. In these examples, stochastic
predictions disagree strongly around locally ambiguous regions and the
second planning pass substantially modifies the selected route. These
examples complement the aggregate statistics by illustrating that the
benefit of reconsideration is concentrated in difficult planning cases
rather than uniformly distributed across the test set.

% ------------------------------------------------------------
\subsection{Visual Re-observation Ablation}
\label{subsec:visual_ablation}

We additionally investigate whether supplying extra visual detail during
reconsideration improves the refined trajectory. This ablation was
performed before the final operating point was fixed, on the 143 queries
triggered at the stricter threshold $\tau=0.8$; the comparison among the
crop variants is internal to this subset and does not depend on the
choice of $\tau$ elsewhere in the paper.

Reconsideration without an additional image achieves a mean ADE of
144.6~px. A broad crop covering the uncertain region yields 150.3~px,
whereas a tightly localized peak-uncertainty crop reduces the mean ADE
to 138.0~px. Although the peak-local crop produces the lowest mean error,
neither crop variant is significantly different from reconsideration
without an additional crop under the paired Wilcoxon tests.

These results suggest that localized visual re-observation can provide
useful additional terrain detail, while broad re-observation does not
necessarily improve the decision. We therefore treat crop design as a
secondary component whose benefit is most apparent when additional
visual information is concentrated around the region of greatest
predictive disagreement.

% ------------------------------------------------------------
\subsection{Comparison with the Physically Executed UGV Trajectory}
\label{subsec:physical_ugv}

We finally compare the underlying VLM planner with the trajectory
physically executed by the UGV. This experiment uses PathGuide~1, which
corresponds to the actual forward route followed by the UGV during data
collection. The remaining four PathGuides associated with each UAV frame
define alternative feasible navigation tasks toward different target
locations.

None of the queries that exceeded the uncertainty threshold originated
from PathGuide~1. Consequently, the adaptive reconsideration branch is
not evaluated by this physical-path experiment. The comparison therefore
evaluates the underlying VLM planning representation rather than
providing a closed-loop validation of the complete UDAV policy.

This distinction is consistent with the qualitative behavior observed
in Fig.~\ref{fig:refinement_result}. Many high-uncertainty cases arise
when the target requires a substantial departure from the nominal
forward direction, such as a lateral maneuver, T-turn, or return toward
a target located behind the vehicle. Such maneuvers are not represented
by the physically executed forward trajectory in PathGuide~1.

Table~\ref{tab:actual_ugv_comparison} compares the human annotation, the
Qwen planner, and SegFormer+A$^{*}$ against the physically executed UGV
trajectory. On the combined data, Qwen achieves an ADE of
131.2~px, compared with 124.2~px for the human PathGuide and 146.7~px
for SegFormer+A$^{*}$. On D2, Qwen obtains the lowest ADE and Chamfer
distance among the three planning representations.

These results indicate that the VLM-generated trajectories remain
geometrically consistent with physically feasible off-road motion,
despite being produced directly from aerial imagery rather than through
an explicitly constructed semantic cost map. A$^{*}$ remains
substantially faster computationally; the purpose of this comparison is
therefore not to demonstrate computational superiority, but to assess
how closely the different planning representations correspond to the
trajectory that was physically executed by the UGV.

\begin{figure}[t]
    \centering
    \includegraphics[width=0.9\linewidth]{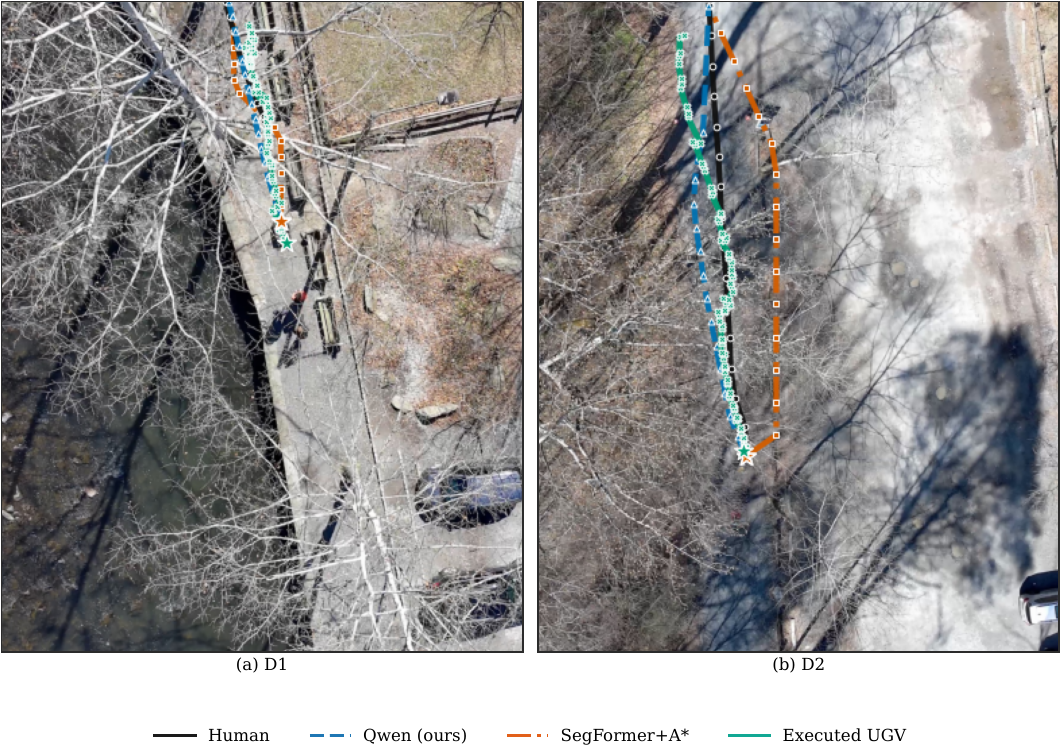}
    \caption{\textbf{Planned trajectories against the path physically
    executed by the UGV.} Two example scenes comparing the human
    PathGuide annotation, the Qwen planner, SegFormer+A$^{*}$, and the
    executed UGV trajectory recovered by registering consecutive UAV
    frames. (a) D1. (b) D2.}
    \label{fig:actual_UGV}
\end{figure}

Representative examples are shown in Fig.~\ref{fig:actual_UGV}.

\begin{table*}[t]
\centering
\caption{Trajectory similarity with respect to the physically executed
UGV path.}
\label{tab:actual_ugv_comparison}
\small
\begin{tabular}{llccc}
\toprule
\textbf{Dataset} &
\textbf{Path Compared with Actual UGV} &
\textbf{ADE (px) $\downarrow$} &
\textbf{Chamfer (px) $\downarrow$} &
\textbf{Hausdorff (px) $\downarrow$} \\
\midrule

D1 & Human
& $\mathbf{134.1 \pm 68.5}$
& $\mathbf{34.4 \pm 14.5}$
& $\mathbf{126.7 \pm 5.0}$ \\

D1 & Qwen
& $145.6 \pm 77.1$
& $41.0 \pm 16.8$
& $143.8 \pm 28.6$ \\

D1 & SegFormer + A*
& $151.4 \pm 72.1$
& $42.7 \pm 9.6$
& $127.4 \pm 5.7$ \\

\midrule

D2 & Human
& $104.4 \pm 51.0$
& $57.9 \pm 16.3$
& $\mathbf{137.0 \pm 58.4}$ \\

D2 & Qwen
& $\mathbf{102.2 \pm 59.3}$
& $\mathbf{48.8 \pm 20.2}$
& $137.5 \pm 68.1$ \\

D2 & SegFormer + A*
& $137.2 \pm 59.5$
& $96.2 \pm 43.9$
& $178.5 \pm 69.1$ \\

\midrule

\textbf{Combined} & \textbf{Human}
& $\mathbf{124.2 \pm 63.9}$
& $\mathbf{42.3 \pm 18.6}$
& $\mathbf{130.1 \pm 33.2}$ \\

\textbf{Combined} & \textbf{Qwen}
& $131.2 \pm 73.6$
& $43.6 \pm 18.0$
& $141.7 \pm 44.6$ \\

\textbf{Combined} & \textbf{SegFormer + A*}
& $146.7 \pm 67.5$
& $60.5 \pm 36.3$
& $144.4 \pm 45.9$ \\

\bottomrule
\end{tabular}

\vspace{1mm}
\footnotesize
Lower values indicate greater similarity to the physically executed
UGV trajectory.
\end{table*}

\section{Conclusion}
We presented UDAV, an adaptive VLM waypoint planner that uses repeated
stochastic trajectory predictions for both self-consistency and geometric
uncertainty estimation. Selecting the medoid of $K=5$ stochastic
predictions already provides a substantially stronger nominal trajectory
than a single deterministic VLM generation, reducing mean ADE from
147.4~px to 115.9~px over the complete permanent test set.

The same samples also yield a normalized geometric disagreement measure
that escalates uncertain queries to a second planning pass. At
$\tau=0.6$, the complete policy achieves a mean ADE of 110.4~px, a
25.1\% reduction relative to deterministic planning, with valid
trajectories for all 400 evaluated queries. The benefit is concentrated
in the high-error tail: UDAV attains the lowest P90 and P95 errors of
all evaluated configurations, below even a $K=10$ consensus baseline, at
an average budget of 7.2 rather than 10 sampled trajectories per query.
On the triggered queries themselves the paired improvement is not
statistically significant, so the second pass acts as a selective
failure-mitigation mechanism, not a guaranteed improvement for every
uncertain query.

We also compared the underlying VLM trajectories with human annotations,
a SegFormer+A$^{*}$ planner, and the trajectories physically executed by
the UGV. Off-road scenes admit multiple feasible routes, and similarity
to a single human annotation is not the only meaningful notion of
planning quality. Because the reconsideration trigger did not activate
on the physically executed PathGuide~1 routes, this experiment is
contextual rather than a closed-loop evaluation of the adaptive policy.

The evaluation has two main limitations: only two flights from a single dataset and environment type were
covered, and the reconsidered trajectories were not executed in closed
loop. Future work will evaluate our proposed planner across more diverse terrain, and deploy the complete policy in
closed-loop UGV execution.

\bibliographystyle{ieeetr}
\bibliography{references}
\end{document}